\pdfoutput=1

\documentclass[11pt]{article}

\usepackage{ACL2023}

\usepackage{times}
\usepackage{latexsym}

\usepackage[T1]{fontenc}

\usepackage[utf8]{inputenc}

\usepackage{microtype}

\usepackage{inconsolata}

\usepackage{bbding}
\usepackage{xcolor}
\usepackage{multirow}
\usepackage{pifont}
\usepackage{graphicx} 
\usepackage{float} 
\usepackage{subfigure} 
\usepackage{booktabs}
\usepackage{colortbl}
\usepackage[export]{adjustbox}
\usepackage{caption}
\usepackage{array}
\usepackage{amsmath}
\usepackage{ragged2e}
\usepackage{amssymb}
\usepackage{adjustbox}
\usepackage{color}
\usepackage{tabularray}
\usepackage{subcaption}

\usepackage{tabularx}
\usepackage{float}
\usepackage{pythonhighlight}

\newcommand{\ourmethod}{\textsc{c-ICL}}

\usepackage{xspace}
\usepackage{graphicx}
\usepackage{amsmath}
\usepackage{tabularx}
\usepackage{enumitem}
\usepackage{mdframed}
\usepackage{xcolor,colortbl}
\usepackage{url}
\usepackage{booktabs}

\newmdenv[
  backgroundcolor=red!05,
  linecolor=quoteborder,
  skipabove=1em,
  skipbelow=0em,
  leftline=true,
  topline=false,
  bottomline=false,
  rightline=false,
  linecolor=red!66,
  linewidth=4pt
]{githubquote}

\newcommand{\tabincell}[2]{\begin{tabular}{@{}#1@{}}#2\end{tabular}}

\title{\ourmethod{}: Contrastive In-context Learning for Information Extraction}

\author{
  Ying Mo\textsuperscript{\rm 1},
  Jiahao Liu\textsuperscript{\rm 2},
  Jian Yang\textsuperscript{\rm 1*}, 
  Qifan Wang\textsuperscript{\rm 3}
  Shun Zhang\textsuperscript{\rm 1}, \\
  \textbf{Jingang Wang}\textsuperscript{\rm 2}, 
  \textbf{Zhoujun Li}\textsuperscript{\rm 1 \thanks{\llap{}\:\: Corresponding author.}}\\
  \textsuperscript{\rm 1}State Key Lab of Software Development Environment, Beihang University, Beijing, China \\  
  \textsuperscript{\rm 2}Meituan, Beijing, China \\
  \textsuperscript{\rm 3}Meta AI, New York, United States
  \{moying, jiaya, shunzhang, lizj\}@buaa.edu.cn, \\ 
  \{liujiahao12, wangjingang02\}@meituan.com, wqfcr@fb.com
}

\begin{document}
\maketitle
\begin{abstract}
There has been increasing interest in exploring the capabilities of advanced large language models (LLMs) in the field of information extraction (IE), specifically focusing on tasks related to named entity recognition (NER) and relation extraction (RE).
Although researchers are exploring the use of few-shot information extraction through in-context learning with LLMs, they tend to focus only on using correct or positive examples for demonstration, neglecting the potential value of incorporating incorrect or negative examples into the learning process.
In this paper, we present \ourmethod{}, a novel few-shot technique that leverages both correct and incorrect sample constructions to create in-context learning demonstrations. This approach enhances the ability of LLMs to extract entities and relations by utilizing prompts that incorporate not only the positive samples but also the reasoning behind them. This method allows for the identification and correction of potential interface errors.
Specifically, our proposed method taps into the inherent contextual information and valuable information in hard negative samples and the nearest positive neighbors to the test and then applies the in-context learning demonstrations based on LLMs. 
Our experiments on various datasets indicate that \ourmethod{} outperforms previous few-shot in-context learning methods, delivering substantial enhancements in performance across a broad spectrum of related tasks. 
These improvements are noteworthy, showcasing the versatility of our approach in miscellaneous scenarios.
\end{abstract}

\section{Introduction}
Information Extraction (IE) is an important task in natural language processing, which aims to obtain structured knowledge from plain text. It can be applied to different domains, such as knowledge graph construction \cite{KG-survey-2023} and question answering \cite{QA-2006}. 
With the rise of large language models \cite{gpt3-2020,ICL-2022,llama-2,gpt4-2023}, information extraction has made significant progress \cite{gpt-ie-li-2023,llm-ie-survey-2023}.
\begin{figure}[t]
\begin{center}
    \includegraphics[width=0.95\columnwidth]{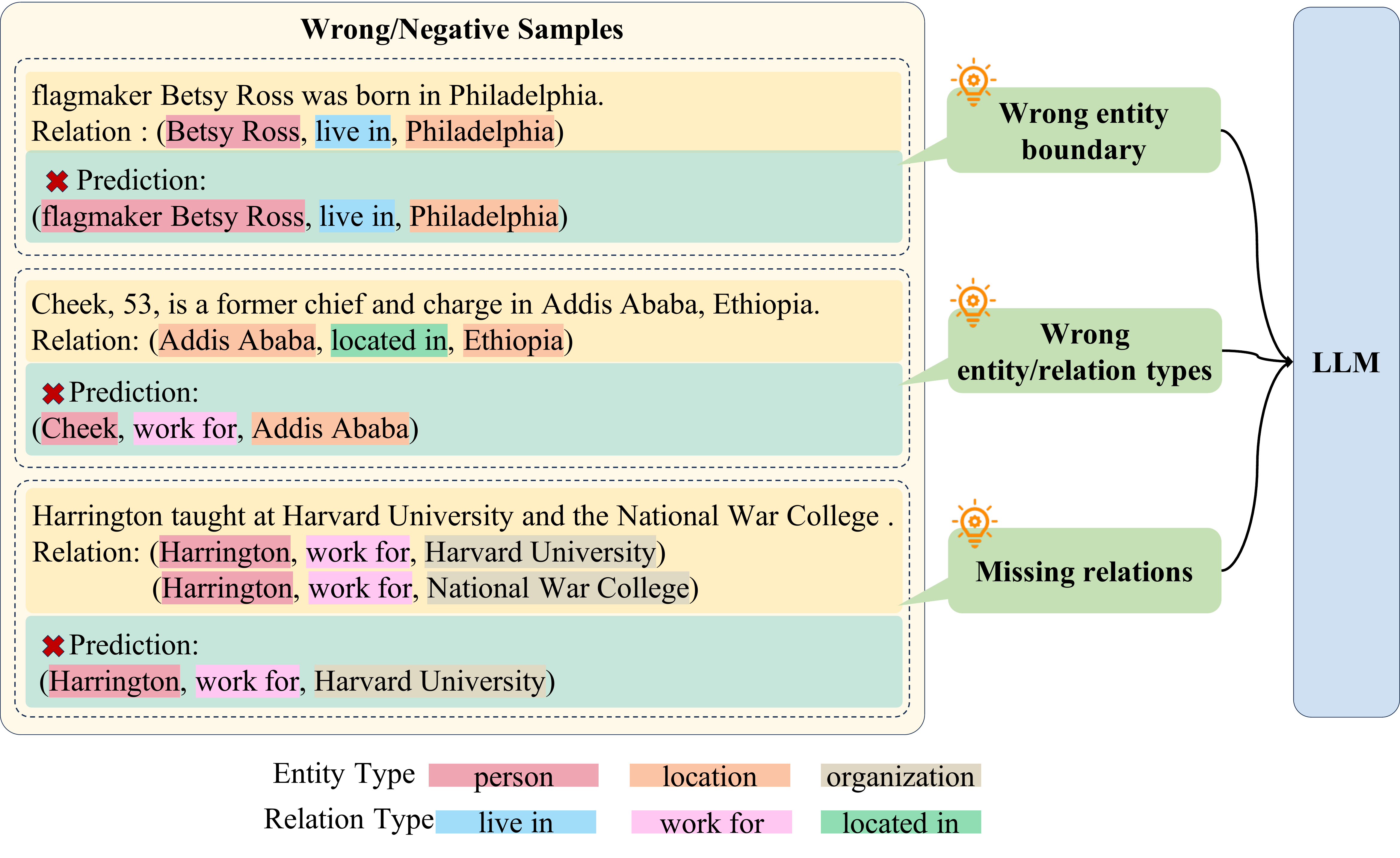}
    \vspace{-2mm}
    \caption{Illustration of our method \ourmethod{} using wrong/negative samples. Take the RE task as an example to illustrate. Wrong/negative samples possess valuable information that LLMs may use to predict the type of IE errors, prompting the model to avoid or correct similar errors.}  
    \label{Fig.intro} 
    \vspace{-6mm}
\end{center}
\end{figure}

Recent advancements in few-shot IE have shifted focus from traditional supervised fine-tuning approaches towards in-context learning (ICL) demonstrations with large language models (LLMs) \cite{ICL-NER-2023,Z-ICL-2023}. 
Prior works \cite{ChatIE-2023, ICL-NER-2023,CoT-ER-2023,Fine-Tuned+Flan-T5+CoT} have explored the use of natural language prompts or ICL demonstrations to guide LLMs in labeling test data under few-shot settings, sometimes requiring additional pre-training or fine-tuning steps. 
To align more closely with the structured nature of information extraction tasks, recent methods \cite{CodeIE-2023,GoLLIE-2023,InstructUIE-2023,Code4Struct-2023,GPT-RE-23} adopt code-like or structured prompts to improve the consistency between pre-training and inference. \textit{However, these methods can not fully unleash the potential of LLMs, partly due to the reliance of models on limited positive data and their inability to learn from their own errors.}
To address this, we propose a contrastive in-context learning approach that utilizes both correct (positive) and incorrect (negative) examples to enhance the learning process of LLMs by exposing them to a broader spectrum of scenarios, including typical mistakes. This method aims to exploit the often-overlooked value in negative data, thus enabling more comprehensive and robust information extraction capabilities.
Assume that the model has learned its own tasks and problem-solving modes from the positive IE data set, but the prediction is still wrong. Should it think about the reasons for the errors, summarize the types of reasons, and try to avoid the above problems in the subsequent inference? Then, adding negative sample-related information can help solve this problem. Inspired by this, we integrate right/positive and wrong/negative examples as ICL demonstrations to enhance the performance of in-context learning IE. 
Specifically, we first use a large model to generate the label of annotated data to select hard negative samples, then select positive samples semantically similar to the current test data from training data, and then design the most in-context demonstrations using different models (NL-LLMs or Code-LLMs ). In the module that selects wrong/negative samples that contain more knowledge, we use semantic similarity-aware self-consistency to conduct ranking.

To demonstrate the superiority of our method, we conduct experiments on three NER and four RE benchmarks and carefully analyze the benefits of our approach. 
Our main contributions are summarized as follows: 
\begin{itemize}
    \item We develop contrastive in-context learning with both right/positive and wrong/negative instances demonstrations, simultaneously enhancing LLMs to extract valuable knowledge for information extraction.
    \item We select hard negative samples based on the effective retrieval strategy as in-context learning, which leverages to enhance the ability of information extraction.
    \item We conduct comprehensive experiments on benchmarks to demonstrate the performance of the proposed method, establishing new state-of-the-art results on most datasets.
\end{itemize}
\section{Task Formulation}
Given a sentence $X$ with $l$ tokens $x_1, x_2, \cdots, x_l$, IE tasks are to predict structured target $Y$ ( NER or RE in this paper) from $x$.
The target $Y$ in NER is entity spans $E((e,t)|x_i,\dots,x_j)$ with entity types. the $e$ is entity in the sequence, and $t$ is the entity type in the predetermined entity types $\mathcal{T}$ (such as \texttt{LOC}, \texttt{PER}, \texttt{ORG}). 
In the RE task, the target $Y$ is a set of relations within entities, usually expressed in the form of a triple $(e_1, r, e_2)$. 
We not only predict the $r \in \mathcal{R}$, but also the entity types $t_1$, $t_2$ of  $e_1$, $e_2$. $\mathcal{R}$ denotes the relation types (e.g., \texttt{Work For}, \texttt{Live In}, \texttt{Located In}). types of $e_1$, $e_2$ also should be predicted. $t \in \mathcal{T}$ means the entity type. 
We treat IE as a text generation task that completes the text to be predicted targets via LLMs. 
We define the task name as type instruction $\mathcal{I} $ to prompt the LLMs for NER or RE. When in NL-LLMs, its representation is a text sequence. In code-style prompts, it is treated as comments inspired by \cite{CodeIE-2023}. 
We transform few-shot sample demonstrations into three parts: the first part is the sentence text, the second is the targets, and the third is a flag denoting whether the sample is positive or not. In NL-LLMs, the output is a list of tuples $[(e_1, t_1), \dots,(e_j,t_j)]$. while in code generation, It is expressed as the operation of adding tuples to the dictionary like `` entity\_dict[``person''].append[``Steve''] ''.
The given test sentence has the first two parts, like the few-shot sample demonstrations. In our work, we pay more attention to code-style large model information extraction beacause of its structure.
\begin{figure*}[ht]
\begin{center}
    \includegraphics[width=0.95\linewidth]{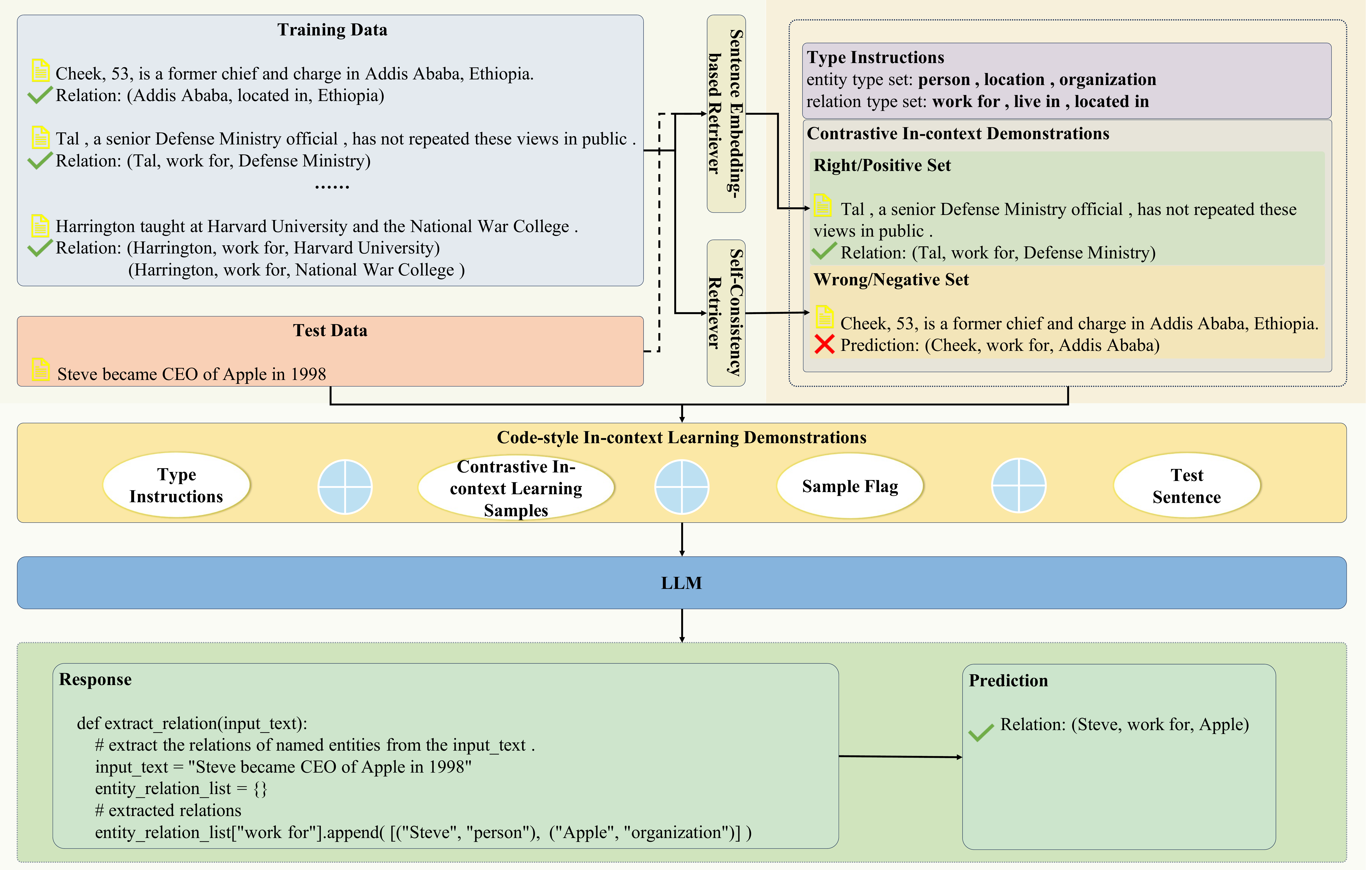}
    \vspace{-1mm}
    \caption{The overview of our method \ourmethod{} for IE. Due to the structures of NER and RE tasks being similar, Take the RE task as an illustration to display the design in this figure. The predictions of the Wrong/Negative set are obtained through LLM. Right/Positive set and Wrong/Negative set are from the training or validation dataset.}  
    \label{Fig.frame_main} 
    \vspace{-5mm}
\end{center}
\end{figure*}
\section{\ourmethod{}}
\subsection{Model Overview}
We introduce \ourmethod{} as shown in Figure \ref{Fig.frame_main}, a novelty few-shot in-context learning method for information extraction, which predicts the right label via contrastive samples using LLMs. 
Unlike prior methods, which only use the samples with gold labels as the in-context learning demonstrations and ignore the knowledge in the wrong/negative samples predicted by models.
Following the prior approaches \cite{CodeIE-2023,Code4UIE-2023}, our \ourmethod{} uses code-style demonstrations for information extraction. 
It consists of building the contrastive samples (right/positive and wrong/negative samples) through retrieval strategies and prompts the LLMs' possible incorrect prediction problems.
\subsection{LLM-based Information Extraction}
Considering the natural language and structure of information extraction, we mainly choose code paragram style LLM to solve the problem. We take the query text as a variable and define the target of the IE task as a variable to return in the functions of the code program, which can illustrate the goal of IE functions. 
Beacause the given defined entity types and relation types have certainty and uniqueness, we think of it as a dictionary library. The type is the keyword in this dictionary, and the entity is the list element under the keyword. 
\begin{equation}
\small
\begin{aligned}
     y_{ner}= \,&\text{entity\_dict[t].append(e)} \\
     y_{re}= \,&\text{entity\_relation\_dict[r].append(}\\
     &\text{[(e1, t1), (e2, t2)])}
\end{aligned}
\end{equation}
where $y$ stands the expected output of the target $Y$ in code-style LLMs. 
$entity\_dict$ and $entity\_relation\_dict$ are the return variables denoting the representations of the different target $Y$ of NER and RE. 
$e$ is the entity span that contains the tokens $x_i,\dots,x_j$ in the given sentence $X$. 
$t$ is one of the entity types $\mathcal{T}$ and $r$ is one of the relation types $\mathcal{R}$.
$t$, $r$ denote the keys of the dictionary in the code function. 
$(e1,t1)$ and $(e2,t2)$ are the subject and object entities in a relation.
\subsection{In-context Demonstrations Construction}
We construct in-context demonstrations for each test sentence fed to the LLMs. Each demonstration consists of the following components.

\begin{figure}[t]
\begin{center}
    \includegraphics[width=1.0\columnwidth]{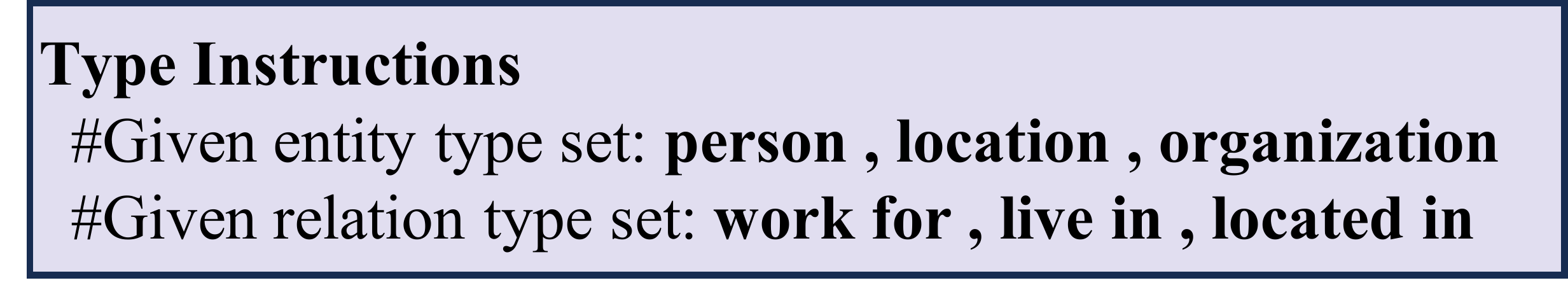}
    \vspace{-5mm}
    \caption{An example of types instructions in-context demonstrations. Take the RE task as an illustration. }  
    \label{Fig.type_icl} 
    \vspace{-5mm}
\end{center}
\end{figure}

\begin{figure}[t]
\begin{center}
    \includegraphics[width=1.0\columnwidth]{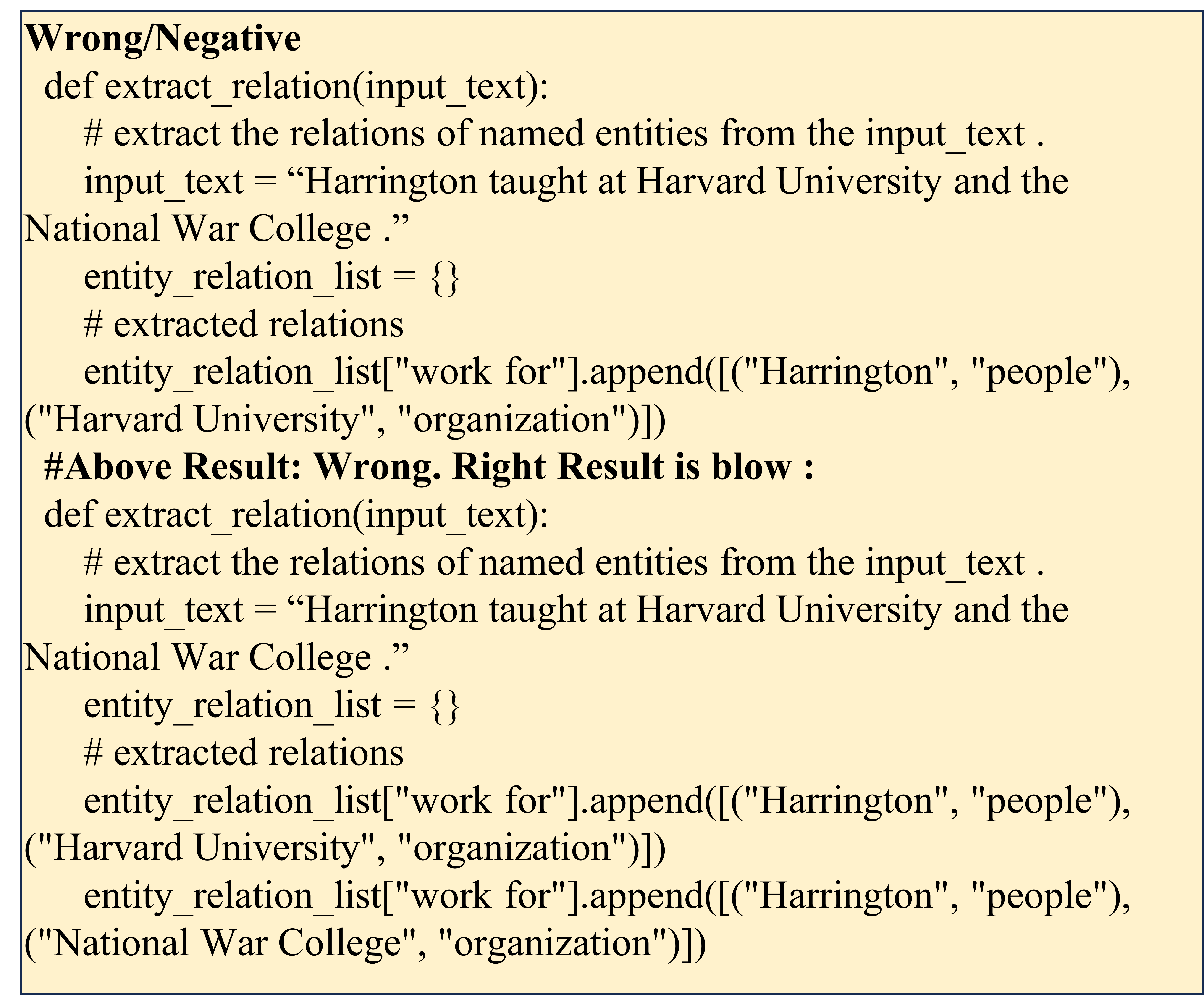}
    \vspace{-7mm}
    \caption{An example of wrong/negative in-context demonstrations. Take the RE task as an illustration. }  
    \label{Fig.wrong_icl} 
    \vspace{-5mm}
\end{center}
\end{figure}

\begin{figure}[t]
\begin{center}
    \includegraphics[width=1.0\columnwidth]{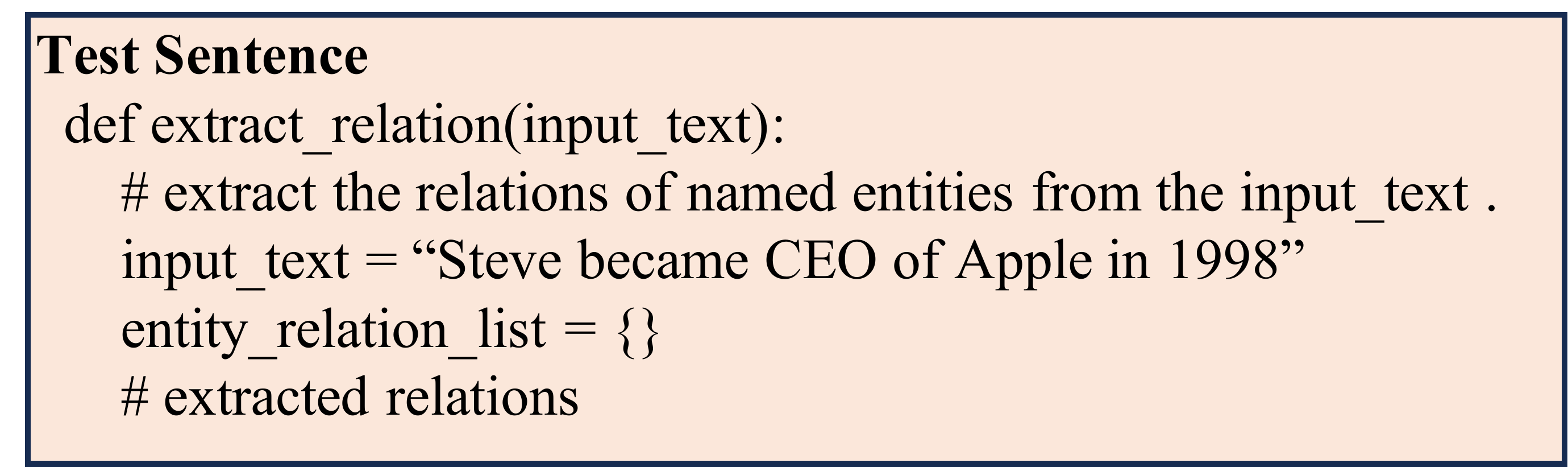}
    \caption{An example of test sentence in-context demonstrations. Take the RE task as an illustration.}  
    \label{Fig.test_icl} 
    \vspace{-5mm}
\end{center}
\end{figure}

\paragraph{Types Instructions} To enhance the model's recognition of types in IE tasks, we provide predefined types ($\mathcal{R}$ or $\mathcal{T}$) in the demonstrations, which prompt LLMs for the purpose of the task, extracting information in a more targeted manner.
We use comments to list possible predefined types instead of using code initialization to represent all types like \cite{Code4UIE-2023,Code4Struct-2023}. If there are too many types, this representation will increase the length of the in-context learning display. We put it at the front of the display to give a hint. 
In our method, The type instructions are shown the following way, illustrated as Figure \ref{Fig.type_icl}:
\begin{equation}
\small
    \mathcal{I_T} = \# \text{Given entity type set: } \mathcal{T} \, .
\end{equation}
\begin{equation}
\small
\begin{aligned}
     \mathcal{I_R} = &\#\text{ Given entity type set: }\mathcal{T} ;\\
    &\# \text{Given relation type set: }\mathcal{R} \, .      
\end{aligned}
\end{equation}where $\mathcal{D_T}$ means the entity type demonstration for NER task. $\mathcal{D_R}$ denotes the relation type demonstration for the RE task. $\mathcal{T}$ likes ``\texttt{LOC}, \texttt{PER}, \texttt{ORG}'' and $\mathcal{R}$ is similar ``\texttt{Work For}, \texttt{Live In}, \texttt{Located In}''.
\begin{figure}[t]
\begin{center}
    \includegraphics[width=1.0\columnwidth]{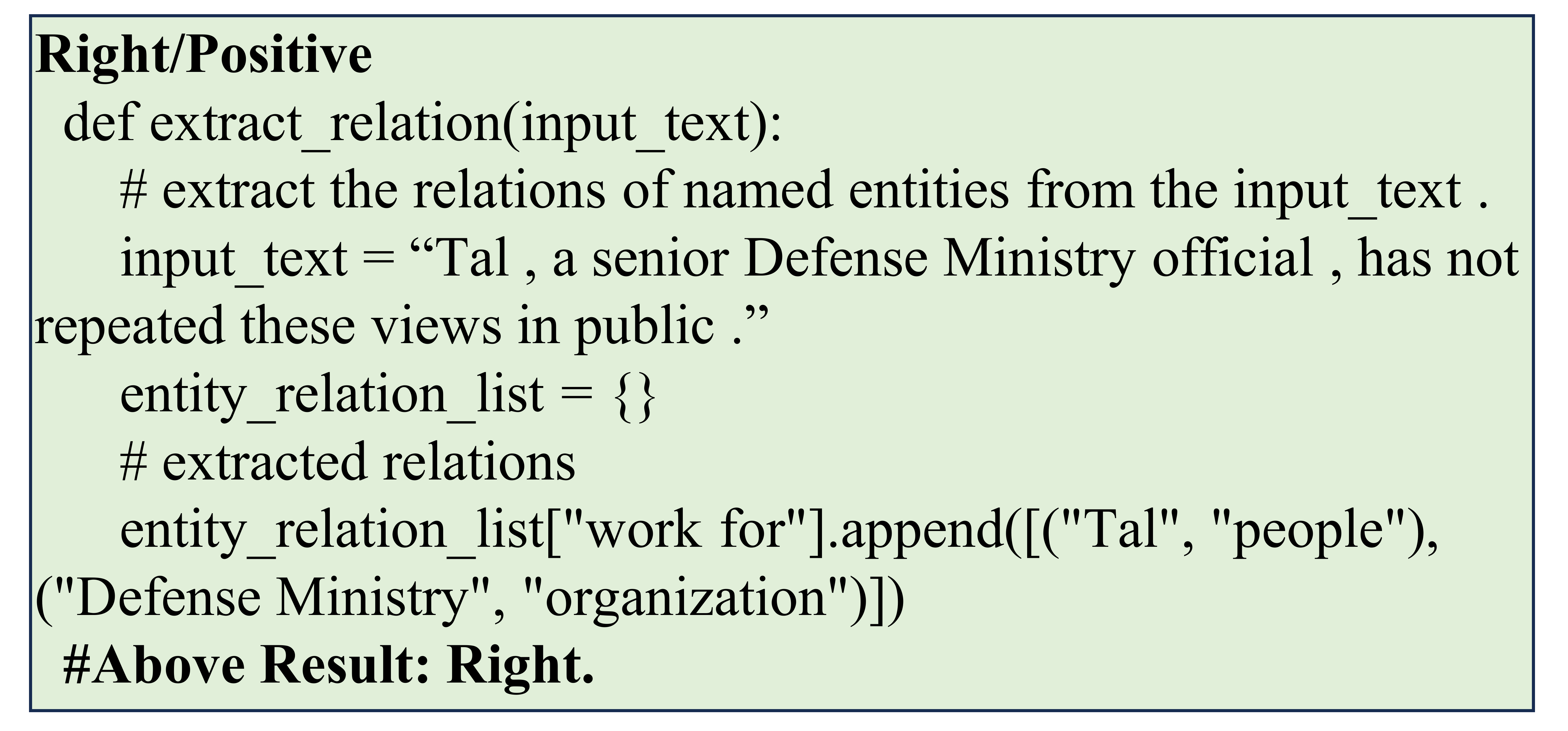}
    \vspace{-7mm}
    \caption{An example of right/positive in-context demonstrations. Take the RE task as an illustration. }  
    \label{Fig.right_icl} 
    \vspace{-5mm}
\end{center}
\end{figure}
\paragraph{Contrastive Samples} In this part, we introduce samples with two essences, one with golden labels and one with wrong labels. 
Positive samples (golden labels) can prompt the large model to generate text with what characteristics. 
Negative samples (wrong labels) are like a set of wrong questions, prompting problems that may occur in the model's inference process and avoiding them. 
The contrastive samples can work together to improve the ability for information extraction.
We choose samples $\{(\hat{X_i},\hat{Y_i})\}_{i=1}^{\hat{n}}$ ($\hat{n}$ is the number of right/positive samples) close to the current test data through semantic similarity and represent them as samples with golden labels. 
Using the self-consistency method, we select hard negative samples ${(\check{X_i},\check{Y_i})}_{i=1}^{\check{n}}$ ($\check{n}$ means the count of wrong/negative samples) as samples with incorrect prediction results. 
Simultaneously, to prompt the large model to correct its errors when the prediction is wrong, we add the instructions (including the flag of whether the prediction is correct and the correct result) to the wrong/negative sample demonstration, as shown in Figure \ref{Fig.right_icl} and \ref{Fig.wrong_icl}.
The contrastive in-context demonstrations are the following:
\begin{equation}
\small
\begin{aligned}
    &\mathcal{\hat{D}} = (\hat{X_1} \oplus \hat{Y_1})\oplus \dots \oplus (\hat{X_{\hat{n}}} \oplus \hat{Y_{\hat{n}}}) \\
    &\mathcal{\check{D}} = (\check{X_1} \oplus \check{Y_1})\oplus \dots \oplus (\check{X_{\check{n}}} \oplus \check{Y_{\check{n}}}) \\
    &\mathcal{D} = \mathcal{\hat{D}} \oplus \mathcal{\check{D}}
\end{aligned}
\end{equation}where $\oplus$ means concatenation. $(X_i \oplus Y_i)$ means the presentation of the function of a sample in the code-style prompt.
If the expression of wrong/positive samples is completely consistent with the positive sample, without training or fine-tuning, it is difficult for LLMs to recognize that this is a negative sample and its meaning. 
To better let negative samples guide the model for avoiding errors, we introduce a flag instruction behind the response to differentiate positive and negative samples. 
Furthermore, the flag instruction contains the right outputs for the given sample, as shown in Figure \ref{Fig.right_icl} and \ref{Fig.wrong_icl}. 
\begin{TinyEquation}
\begin{align}
flag =\begin{cases}
\text{``\#Above Result: Right.''} &\text{if } X \, \text{is }+\\
\text{ ``\#Above Result: Wrong. Right Result is blow: D''} &\text{if } X \, \text{is }- 
\end{cases}
\end{align}
\end{TinyEquation}where $flag$ is the demonstration of judging positive and negative samples. 
$\mathrm{D}$ is the correction corrective representation of the incorrect answer. $+$ and $-$ denote positive and negative samples, respectively.
\vspace{-5mm}
\paragraph{Test sentence} Similar to the above demonstration, it only converts the test text shown in Figure \ref{Fig.test_icl} and entity/relation types into the code-style input of the contrastive sample. 
The final contrastive in-context demonstration is formulated as follows:
\begin{equation}
\small
\begin{aligned}
    Y_{test} = \mathcal{M}(Y|\mathcal{I},\mathcal{D}, X_{test})
\end{aligned}
\end{equation}where $X_{test}$ means the code-style transformation of the test sentence. $\mathcal{M}$ is the large language model like CodeLlama\cite{llama-2}. The RE output of $Y_{test}$ is like a response module in Figure \ref{Fig.frame_main}, and the NER task is similar.
\subsection{In-context Example Retrieval Strategies}
\paragraph{Sentence Embedding-based Retrieval Strategy}
\citet{KATE-knn-2022} indicates that in-context learning demonstrations similar to the test data in semantic embedding may result in more reliable outcomes. 
So, the selection of samples is crucial for few-shot information extraction. 
In this part, we use the sentence embedding-based retrieve to select the samples from the training dataset for in-context learning demonstrations. 
Inspired by the previous works \cite{ICL-Think-2022,KATE-knn-2022,Code4UIE-2023}, we use the \textit{k} nearest neighbors to retrieve sentences. 
After ranking the semantic similarity, we select the top-\textit{k} samples with entities or relations. Our work employs the code LLMs to calculate the semantic similarity via cosine similarity.
\paragraph{Self-Consistency Retrieval Strategy}
For the wrong/negative samples, we select hard negative samples from the training dataset, which contain valuable knowledge and can better reflect the fuzzy performance of large language models in information extraction tasks. We use the large model first to get the prediction results of the training dataset. In this step, we apply self-consistency \cite{self-consistency-2023} with votes to obtain predictions with high confidence. For each prediction, we determine whether the sample is a hard negative sample by calculating F1, which indicates that the prediction is very close to the correct result results of each sample. through this method, we can get high-quality hard negative samples as the wrong/negative samples of contrastive in-context demonstrations.
\section{Experiments}
\subsection{Datasets}
\noindent\textbf{RE Datasets}
For relation extraction, we evaluate on datasets CoNLL04 \cite{conll2004}, ACE05-R \cite{ace05}, NYT \cite{nyt} and SciERC \cite{scierc}. 
We follow \citet{UIE-2022} to split all these datasets. 

\noindent\textbf{NER Datasets}
We evaluate our approach on NER task with CoNLL03 \cite{conll2003}, ACE04\cite{ace04} and ACE05-E\cite{ace05}. and we split the datasets followed by the works \cite{MRC-NER-2020,moying-2023,mcl-ner-moying-2023,CodeIE-2023}.
Table \ref{tab:datasets_anlasis} shows the dataset statistics in Appendix \ref{sec:appendix_a}.

\subsection{Experiments Setting}
For each IE task, we make a k-shot training set following the previous work \cite{CodeIE-2023}, 
which samples k training samples for each entity or relation type via retrieval strategies. 
Since we have introduced wrong/negative samples, the settings are slightly different. 
We try to keep the overall sample number consistent for a fair comparison. 
In our experiments, we set the numbers of samples as 20, 14, and 14 for NER datasets CoNLL03, ACE04, and ACE05-E respectively, introduced in contrastive in-context learning demonstrations.
The numbers of contrastive in-context learning samples for RE datasets CoNLL04, ACE05-R, NYT, and SciERC are 20, 12, 24, and 14.
We use CodeLlama \cite{llama-2} as the backbone and set the max token length to 8k and top\_p to 0.7. 
Setting temperature in [0.3, 0.6, 0.9] is dependent on different datasets. 
When sampling hard negative samples, we query each sample to the model 3 times to acquire a suitable response. 
The experiment details are listed in Appendix \ref{sec:appendix_a}.

\subsection{Evaluation}
As in prior work \cite{UIE-2022, CodeIE-2023}, we use entity F1 score and relation strict F1 score as the evaluation metrics for NER and RE tasks, respectively. 
Note the relation strict F1 score for RE. A relation is correct only if the relation type, entity span, and entity type are all right. 
To ensure the accuracy of the results, we perform three runs with different random seeds for each experiment. 

\subsection{Results}

\begin{table*}[ht]
\begin{adjustbox}{width=0.92\linewidth,center}
\begin{tabular}{lcccccc}
\toprule
\multicolumn{1}{c}{\multirow{2}{*}{Model}} & \multirow{2}{*}{Paradigm} & \multirow{2}{*}{Backbone} & \multicolumn{4}{c}{RE} \\ 
\multicolumn{1}{c}{}                       &                           &                           & CoNLL04              & ACE05                & NYT                  & SciERC               \\ \hline
UIE \cite{UIE-2022}                      & SFT                       & T5-large                  & 75.00                & \textbf{66.06}                & /                    & 36.53                \\
InstructUIE \cite{InstructUIE-2023}      & SFT                       & Flan-T5-11B               & \textbf{78.48}                & /                    & \textbf{90.47}                & \textbf{45.15}                \\ \hline
CodeIE\cite{CodeIE-2023}                  & ICL                      & Code-davinci-002          & 53.10                & 14.02                & 32.17                & 7.74                 \\
Code4UIE \cite{Code4UIE-2023}             & ICL                      & Gpt-3.5-turbo-16k         & 54.40                & 11.50                & 51.70                & /                    \\
CodeKGC \cite{CodeKGC-2023}               & ICL                      & Text-davinci-003          & 49.80                & /                    & /                    & \textbf{24.00}                 \\ \hline
\multirow{3}{*}{\textbf{c-ICL (ours)}}    & ICL                      & CodeLlama-7B             & 53.27                & 18.75                & 58.39                & 12.13                \\
                                        & ICL                        & CodeLlama-13B            & \underline{56.43}                & \underline{20.57}                & \underline{60.16}                & 15.29                \\
                                        & ICL                        & CodeLlama-34B            & \textbf{56.93}                     & \textbf{22.31}                & \textbf{60.92}                     & \underline{17.33}                    \\ \bottomrule
\end{tabular}
\end{adjustbox}
\vspace{-2mm}
\caption{The experiment performances on RE benchmarks. SFT denotes the model adopts supervised fine-tuning with training data. ICL means the model uses in-context learning.}
\vspace{-2mm}
\label{tab:main_results_RE}
\end{table*}
\begin{table*}[ht]
\begin{adjustbox}{width=0.92\linewidth,center}
\begin{tabular}{lcccccl}
\toprule
\multicolumn{1}{c}{\multirow{2}{*}{Model}} & \multirow{2}{*}{Paradigm} & \multirow{2}{*}{Backbone} & \multicolumn{3}{c}{NER}   &  \\ 
\multicolumn{1}{c}{}                       &                           &                           & CoNLL03 & ACE04 & ACE05-E &  \\ \hline
UIE \cite{UIE-2022}                         & SFT                       & T5-large                  & \textbf{92.99}   & \textbf{86.89} & 85.78   &  \\
InstructUIE \cite{InstructUIE-2023}         & SFT                       & Flan-T5-11B               & 92.94   & /     & \textbf{86.66}   &  \\ \hline
CodeIE \cite{CodeIE-2023}                  & ICL                       & Code-davinci-002          & 82.32   & \textbf{55.29} & 54.82   &  \\
Code4UIE \cite{Code4UIE-2023}              & ICL                       & Gpt-3.5-turbo-16k          & 79.70   & 54.0 & \textbf{57.00}   &  \\
Self-Improving\cite{Self-Improving-NER-2023}                           & ICL                       & Gpt-3.5-turbo             & 83.51   & /     & 55.54   &  \\ \hline
\multirow{3}{*}{\textbf{c-ICL (ours)}}     & ICL                       & CodeLlama-7B             & 83.98   & 47.88 & 45.65    &  \\
                                           & ICL                       & CodeLlama-13B            & \underline{85.62}   & 49.69 & 48.04    &  \\
                                           & ICL                       & CodeLlama-34B            & \textbf{87.36}   & \underline{54.47} & \underline{55.65}         & \\ \bottomrule
\end{tabular}
\end{adjustbox}
\vspace{-2mm}
\caption{The experiment performances on NER benchmarks. SFT denotes the model adopts supervised fine-tuning with training data. ICL means the model uses in-context learning. }
\label{tab:main_results_NER}
\vspace{-5mm}
\end{table*}
\textbf{RE Results} Table \ref{tab:main_results_RE} shows the results of the RE task. 
Among the datasets, NYT has the most significant improvement, exceeding CodeIE and Code4UIE by +26.22\% +7.22\%, respectively. 
Our method get an improvement of 4.73\% on the F1 score compared to the LLM-based baselines for ACE05. 
Although CoNLL04 on CodeLlama-7b is slightly weaker than Code4UIE, there is a specific improvement on CodeLlama-13b and  CodeLlama-34b by +2.03 points. 
At SciERC, our method improves by +4.39 points compared to CodeIE but is lower than CodeKGC. 
It is lower than CodeKGC because CodeKGC re-structure the corpus into code format and builds a new dataset for pre-training to effect rationale-enhanced generation. In addition, the backbone of CodeKGC is a larger base model\cite{prompting-reasoning-iclr2023} than that we used.  

\noindent \textbf{NER Results} Table \ref{tab:main_results_NER} shows the results of the NER task. 
Our method achieves superior performance overall compared with the previous baselines, proving the effectiveness of our method in the information extraction subtask NER. 
We get an F1 score of 87.36\% on the CoNLL03 dataset, increased by +3.76 points compared to the in-context learning methods CodeIE \cite{CodeIE-2023}, Code4UIE\cite{Code4UIE-2023}, and Self-Improving \cite{Self-Improving-NER-2023}. 
for ACE05-E, our model performs slightly better than the CodeIE method by 0.54 points. 
For ACE04, our results are weaker than those above LLM-based baselines. 
The improvement in performance of our method is not significant on these two datasets overall.
The main reasons include that 
1) these two datasets contain nested entities, and their error types are more numerous and complex compared to those of common NER tasks;
2) the added wrong/negative samples, which may lead to longer text length, affect the capture of contextual information, and lead to performance degradation;
3) The LLMs-based baselines are the GPT-based\cite{gpt3-2020,openai-2022-chatgpt} methods, which are stronger pre-trained large language model. 
It should be noted that LLMs also have limitations with the nested ACE04 and ACE05-E NER benchmarks during the experiments and enhance reasoning for complex situations. 
To further verify our method, we reproduce CodeIE and replace the backbone with CodeLlama.
See Appendix \ref{sec:appendix_b} for relevant results.

Compared with the supervised baselines UIE\cite{UIE-2022} and InstructUIE \cite{InstructUIE-2023}, our method performs worse than them, but it is approaching them on CoNLL03, showing that our method gives the model an excellent hint to enhance the reasoning ability on this task.

\section{Further Analysis}
\label{analysis}
\subsection{Ablation Study}
\begin{figure}[t]
\begin{center}
    \subfigure{
        \centering
        \includegraphics[width=0.46\columnwidth]{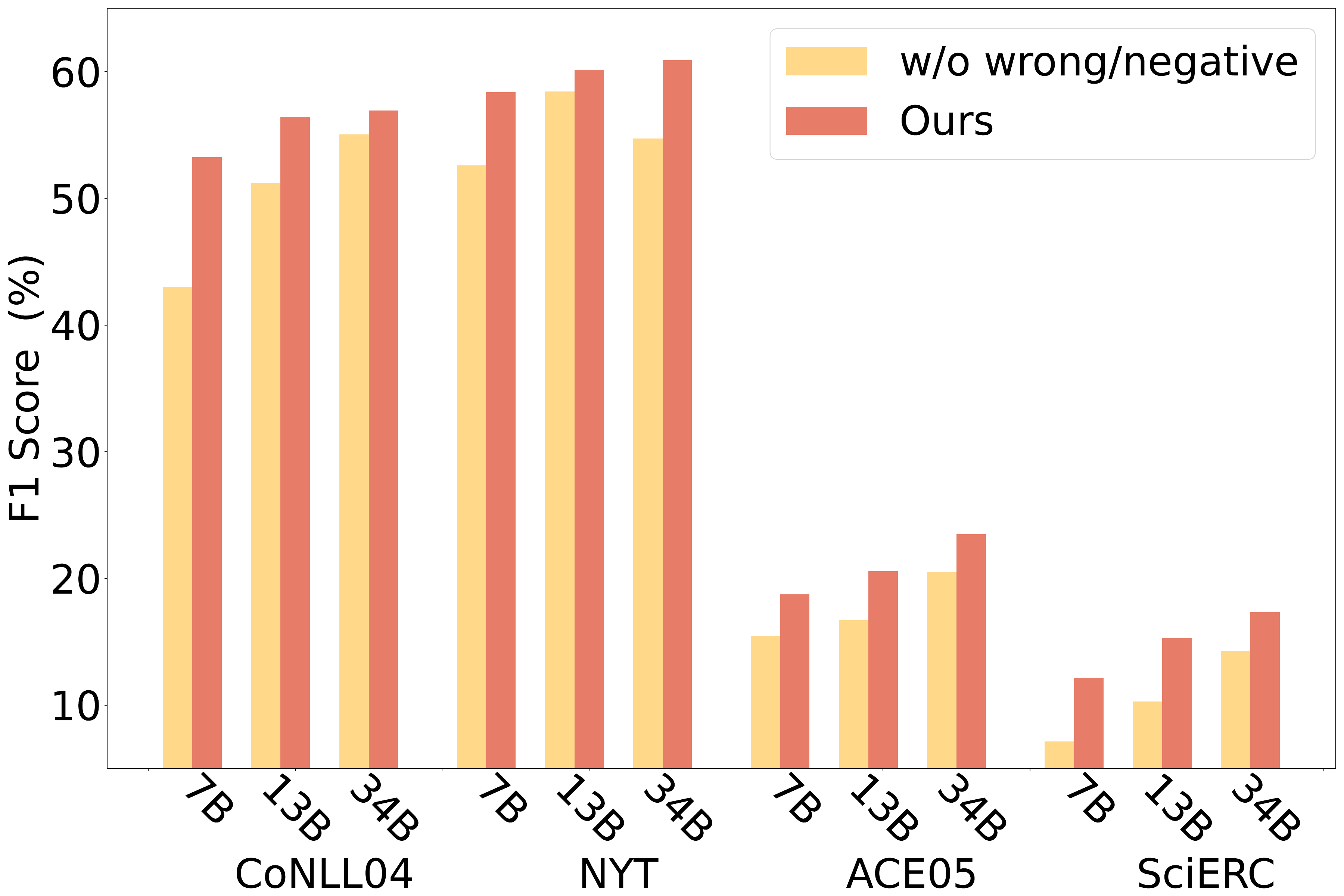}
    }
    \subfigure{
        \centering
        \includegraphics[width=0.46\columnwidth]{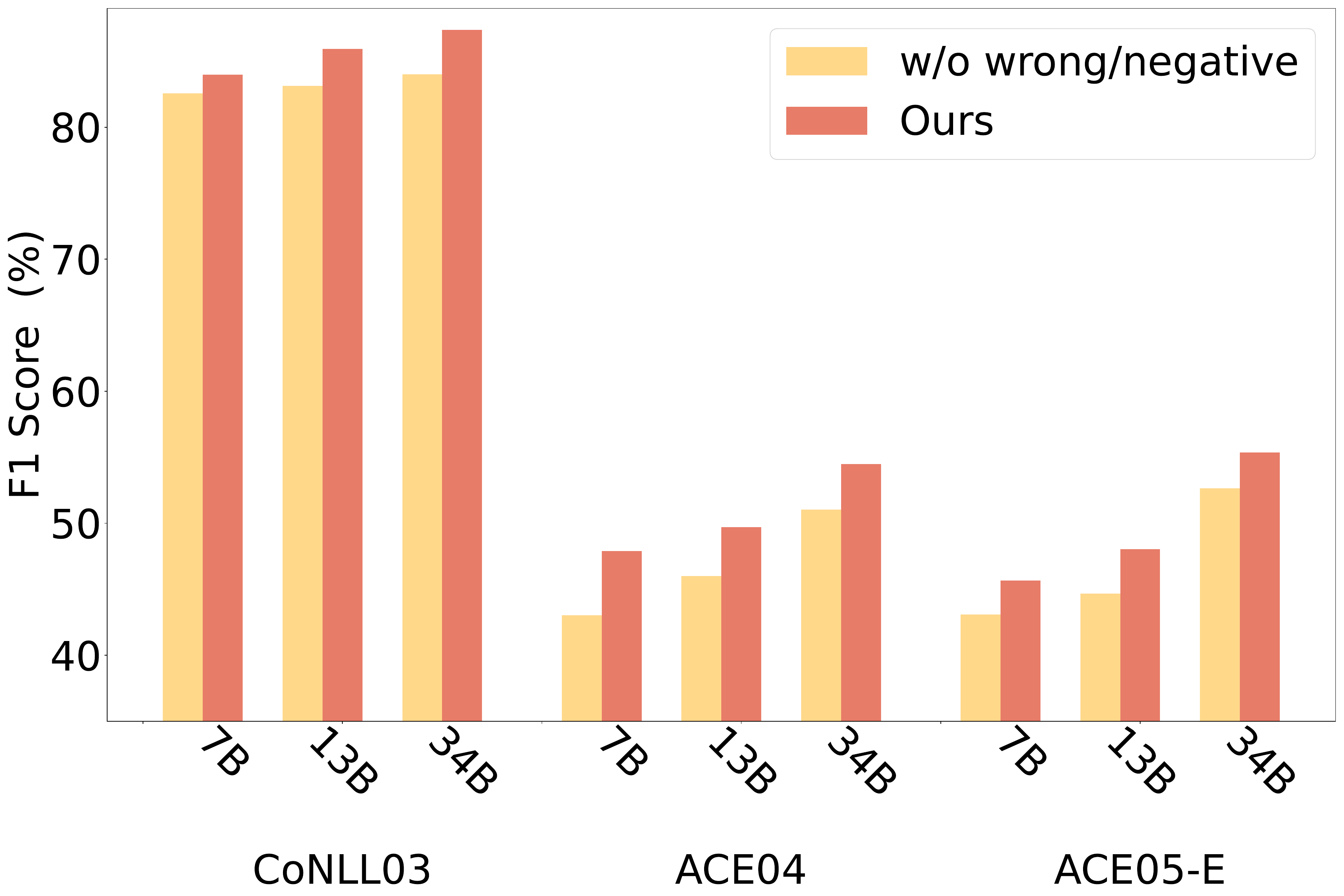}
    }
    \vspace{-3mm}
    \caption{Alation study of \ourmethod{} based on different CodeLlama for IE. 
    Left and right figures denote the ablation study of RE and NER tasks, respectively.}
    \label{Fig.ablation_study} 
    \vspace{-5mm}
\end{center}
\end{figure}

To demonstrate the effectiveness of the proposed method, we conduct an ablation study of our method. We run experiments on RE and NER datasets based on CodeLlama-7b, CodeLlama-13b, and CodeLlama-34b. The results are demonstrated in Figure \ref{Fig.ablation_study}. 
{\large{\ding{172}}} Ours, which is the final approach with the contrastive in-context objectives; 
{\large{\ding{173}}} w/o wrong/negative, which denotes that the method only adopts the right/positive samples as in-context demonstrations. 
From the ablation experiments, our method \ourmethod{} outperforms different levels 
of improvement effects than {\large{\ding{173}}}, indicating that the contrastive in-context learning with right/positive and wrong/negative samples could prompt the LLMs to learn the positive and effective knowledge information. It can be seen that the larger the model, the better the effect for IE tasks.

\subsection{Examples of Contrastive ICL}
\begin{figure*}[ht]
\begin{center}
    \subfigure[]{
        \includegraphics[width=0.6\columnwidth]{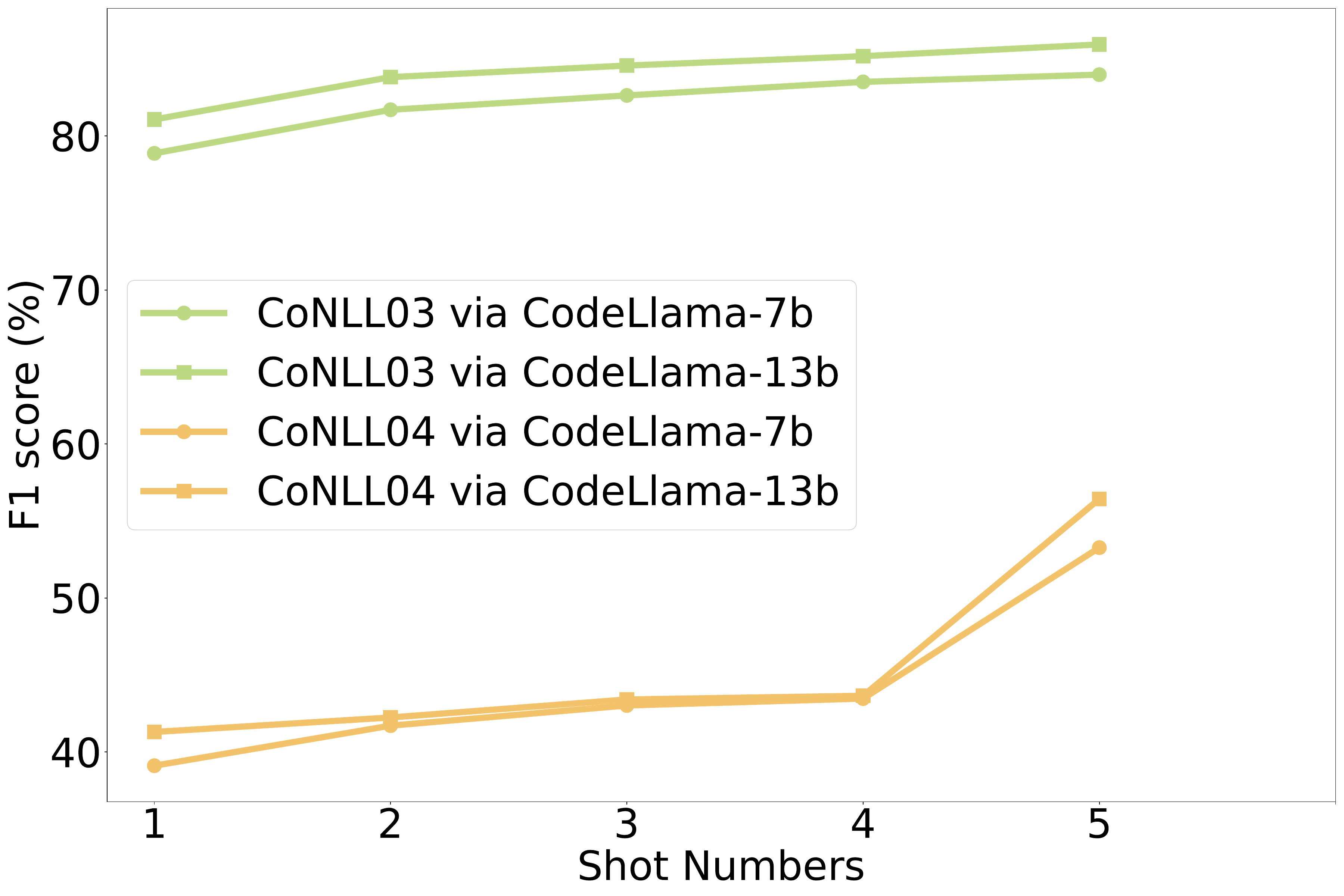}
        \label{Fig.shot_numbers}     
    }
    \subfigure[]{
        \includegraphics[width=0.6\columnwidth]{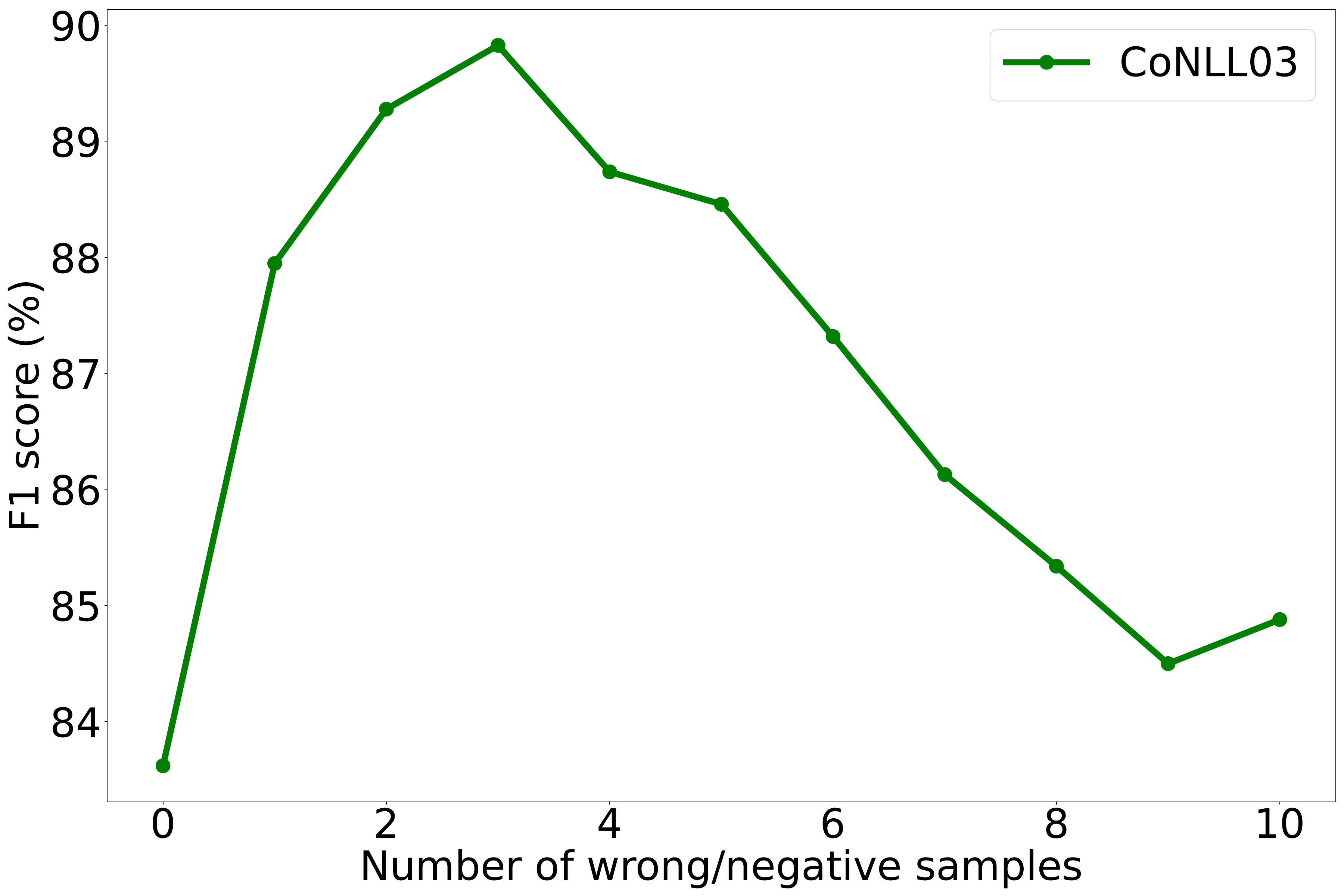}
        \label{Fig.proportion_positive_negative_conll03} 
    }
    \subfigure[]{
        \includegraphics[width=0.6\columnwidth]{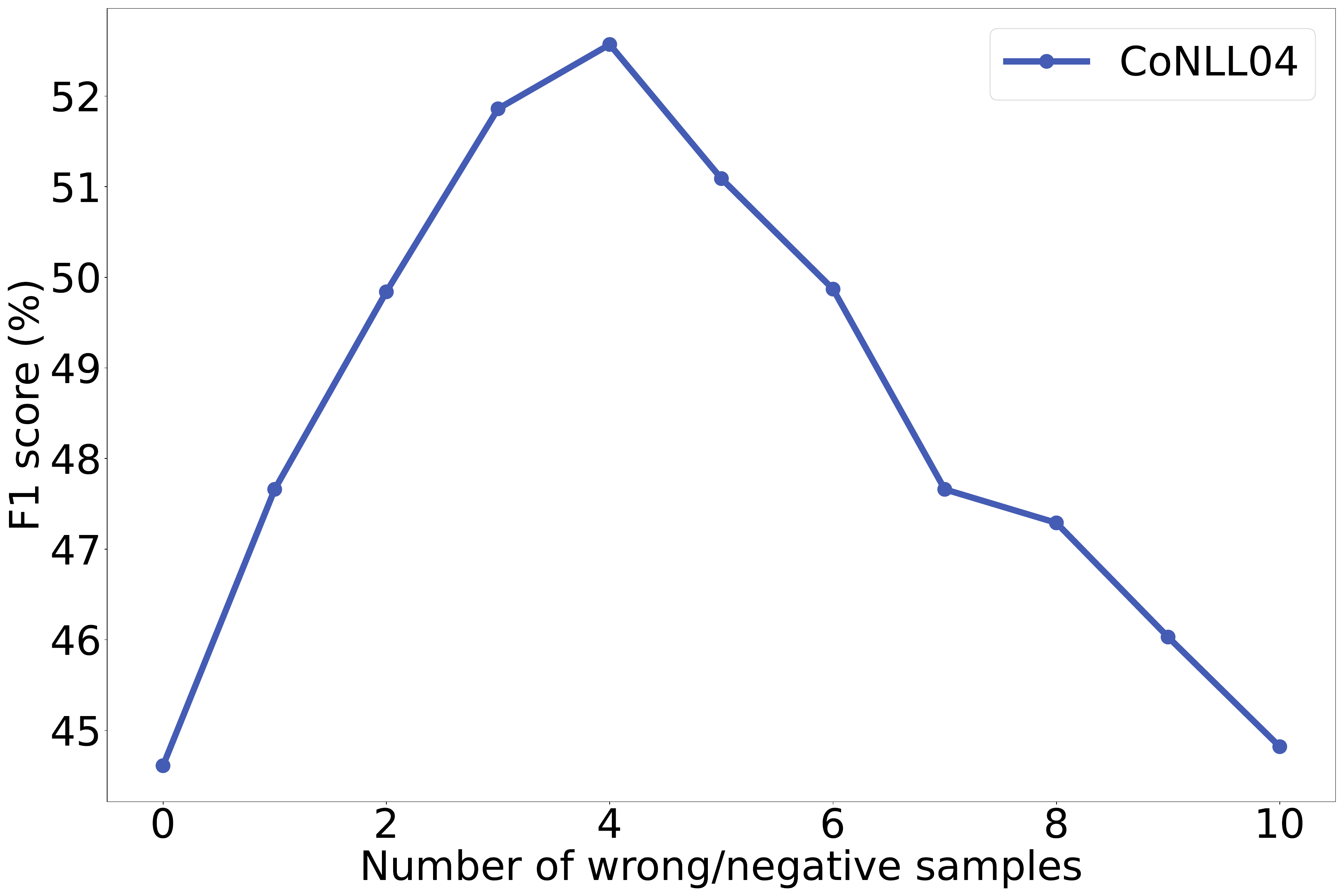}
        \label{Fig.proportion_positive_negative_conll04} 
    }
    \vspace{-3mm}
    \caption{The performance of numbers of contrastive in-context demonstrations. 
    (a) means the results of CoNLL03 (NER) and CoNLL04 (RE) with different shot numbers. 
    (b) and (c) present the results of the proportion of positive and negative samples on CoNLL03 and CoNLL04, where we sample 300 test data for analysis.}
    \label{Fig.sample_num}
    \vspace{-15pt}
\end{center}
\end{figure*}

\paragraph{Different Shot Numbers}
To illustrate the impact of different numbers of contrastive in-context demonstrations on information extraction, we conducted further experiments on the CoNLL03 (NER) and CoNLL04 (RE) via CodeLlama-7b and CodeLlama-13b. 
In this part, we ensure that the number of wrong/negative samples is consistent with two and that other display samples are right/positive. 
The results are presented in Figure \ref{Fig.shot_numbers}. 
As can be seen from it, the effect of each IE task increases as the number of shots increases.
Overall, the effectiveness of each IE task tends to increase with the number of shots. 
This phenomenon is because as the number of sample instances increases, the large language models can glean more in-context information from them.
\begin{figure}[ht]
\begin{center}
    \subfigure{
        \centering
        \includegraphics[width=0.85\columnwidth]{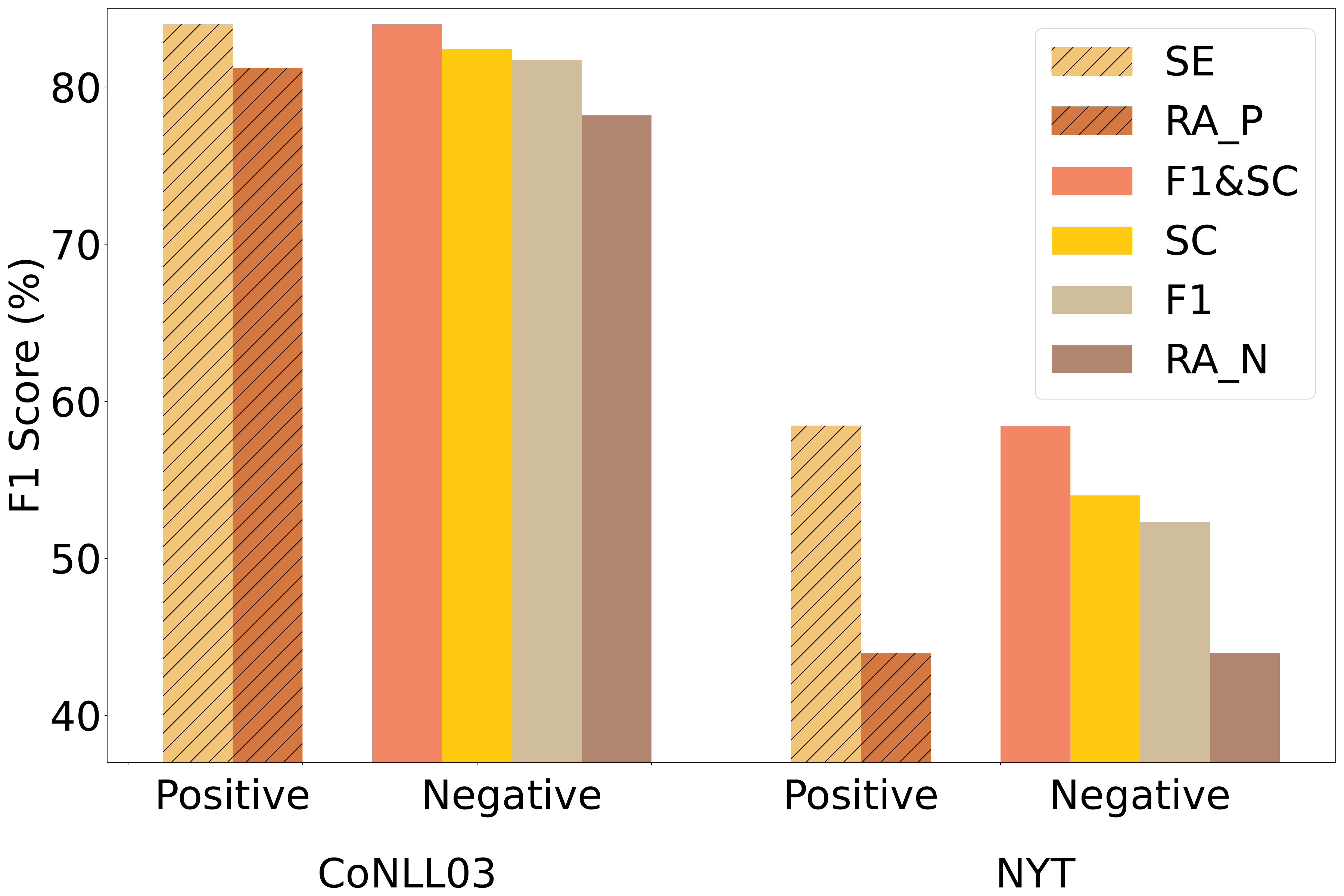}
    }
    \vspace{-3mm}
    \caption{The comparison of retrieval strategies for contrastive samples. The bars with hatched and not show retrieval strategies for positive and negative samples, respectively. 
    SE refers to the sentence embedding-based retrieval strategy, 
    RA\_$*$ to random sample, 
    SC to the self-consistency strategy, 
    F1 to retrieve sample by the F1 score, 
    F1\&SC to retrieve sample via F1 and SC.} 
    \label{Fig.retrieval_strategy} 
    \vspace{-20pt}
\end{center}
\end{figure}
\paragraph{Proportion of Positive and Negative Demonstrations}
We conduct experiments to analyze the proportion of positive and negative samples in contrastive in-context demonstrations. 
In this part, we evaluate the CoNLL03 (NER) and CoNLL04 (RE) datasets and run experiments on the CodeLlama-7b. 
Note that we sample 300 test data for this analysis. 
We set the total number of demonstration samples to a particular value; the total number of contrastive samples is fixed, and the numbers of positive and negative samples are changing. 
The results are illustrated in Figure \ref{Fig.proportion_positive_negative_conll03} and \ref{Fig.proportion_positive_negative_conll04}. 
Overall, the effectiveness of information extraction tends to rise first and then fluctuates with the increase in negative sample numbers. 
When positive samples provide adequate contextual information, adding negative samples can indirectly prompt the model to acquire other knowledge (types of errors that may occur) in entity or relation recognition and modify erroneous predictions. 
However, adding too many wrong/negative samples may increase noise, making the model ambiguous in identifying positive and negative knowledge. 
It will not serve as an optimistic prompt, resulting in poorer results. 
Besides, adding more negative samples makes the text longer in our design, which can lead to worse effects. 
Therefore, preferring a proper number of contrastive in-context demonstrations is necessary.

\subsection{Retrieval Strategy of Contrastive ICL}
We run the experiments of different retrieval strategies of contrastive in-context demonstrations in our method on CoNLL03 and NYT.
The bars hatched in Figure \ref{Fig.retrieval_strategy} show the results for sampling the positive samples from training data. 
The sentence embedding-based retrieval strategy for positive samples performs better than random sampling positive data. 
In this strategy, the samples are more similar to the current test sequence in semantics.
The bars without hatched in Figure \ref{Fig.retrieval_strategy} exhibit the results of different retrieval strategies for the wrong/negative samples. 
We find that 1) combining the self-consistency retrieval strategy and setting an F1 score threshold for retrieving hard negative samples can result in better performance. 
The strategy can obtain high-quality samples to prompt LLMs to learn from more knowledge. 
2) Random retrieval for sampling wrong/negative data has lower effects than other strategies and it would cause fluctuation. 

\begin{figure}[ht]
\begin{center}
    \subfigure{
        \includegraphics[width=1\columnwidth]{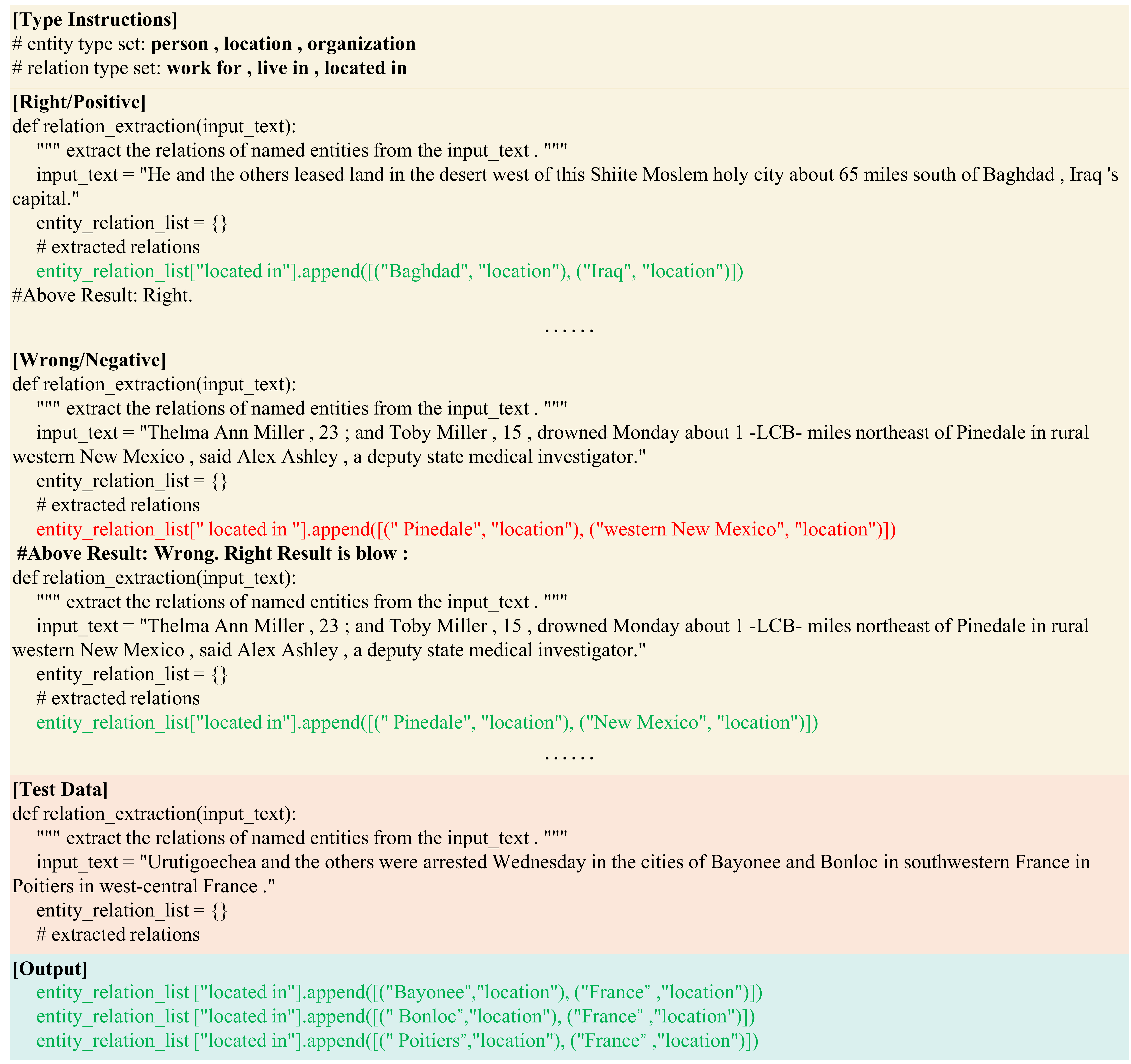}
        \label{Fig.case_study_1}     
    }
    \vspace{-5mm}
    \caption{Case study. 
    The example stands for the results of RE. 
    The red text means the incorrect labels and the green text means the right labels.   
    }
    \label{Fig.case_study}
    \vspace{-6mm}
\end{center}
\end{figure}
\subsection{Case Study}
We select some cases of typical test samples to illustrate better the amendment of our method in Figure \ref{Fig.case_study}. 
given example 1, we show the more similar positive and negative samples as contrastive ICL demonstrations in the RE task.
In this case, ``southwestern France'' and ``west-central France'' in the test sample are easily identified as ``location'' entities. 
The introduction of negative samples in ICL demonstrates potential issues that may arise during the test inference process. 
It further points out errors and provides corrective prompts, assisting the LLM model to accurately infer the correct results. 
We also discuss the case of the NER task. More cases can be seen in Appendix~\ref{sec:appendix_c}.

\section{Related Work}
\paragraph{Generative Information Extraction}
Due to the differences in subtasks (e.g. NER and RE), information extraction (IE) has seen many task-specific supervised models, including understanding models~\cite{Lample2016,tang2018,span-ner-yu2021,wang2018response,wang2020formality,wang2019harnessing,pl-maker-2022} and generation models~\cite{bartner-2021,UIE-2022,Lample2016,tang2018,moying-2023}, span-based models~\cite{span-ner-yu2021,pl-maker-2022}. 
Generative Transformer-based models have been adapted for many NLP tasks, such as information extraction \cite{crop,mcl-ner-moying-2023,zeroshot_IE_ChatGPT} and machine translation \cite{um4,hlt_mt,microsoft_wmt2021,ganlm,LLM_MNMT}. 
Through scaling up the model size, large language models (LLMs) produce competitive results without training by unifying different IE subtasks. 
NL-LLMs~\cite{ChatIE-2023} converts structured information tasks into natural language, but there is an inconsistency between the reasoning goal and pre-training. 
Code-LLMs~\cite{qurg,CodeIE-2023,Code4UIE-2023,CodeKGC-2023} converts the text-to-structure IE task into a structure-to-structure code generation task.
\vspace{-5pt}
\paragraph{In-context Learning}In-context learning (ICL) can be enhanced in large-scale LLMs \cite{gpt3-2020,llama-2,qwen,owl} by constructing valuable demonstrations \cite{learn_to_select,KATE-knn-2022,retrive-ICL-2022,ICL-survey-dong2023,wang2023adaptive}. Some researchers \cite{KATE-knn-2022,Code4UIE-2023} proposes to use \textit{K}NN method to retrieve similar samples and \citet{GPT-RE-23} employs task-aware retrieval and gold label-induced reasoning representation to select appropriate samples.
\citet{ChatIE-2023,sc-two-vote-2023} use dialogue and question-answer methods. 
Code-style prompts convert them into program methods \cite{CodeIE-2023} or classes \cite{Code4UIE-2023,CodeKGC-2023}.
\vspace{-1mm}
\paragraph{Hard negative sample}
Hard negative samples should have different labels from the anchor sample but have embedding features very close to the anchor embedding. Different from learning and transferring knowledge through positive samples, models may obtain valuable information from negative samples to enhance the model performance \cite{cl-hard-2021,filter-hard-2023,mcl-ner-moying-2023}. Introducing negative samples can directly assist with positive samples to comprehensively extract helpful knowledge for LLM.

\vspace{-1mm}
\section{Conclusion}
In this work, we introduce \ourmethod{}, contrastive in-context learning for few-shot information extraction, including right/positive and wrong/negative demonstrations. 
In addition through type instruction demonstrations prompt mention tags in the IE task. 
From the contrastive samples, the LLMs could obtain effective information and indirect but positive, valuable additional knowledge for IE tasks. 
Besides, our method adopts semantic similarity retrieval strategies and self-consistency votes to retrieve in-context examples better suited for the current sentence and task, significantly improving IE performance.
Extensive experiments prove the effectiveness of \ourmethod{} on various benchmarks.
\clearpage
\section*{Limitations}
We acknowledge the following limitations of this study: (1) This work focuses on exploring the in-context learning for few-shot NER and RE tasks. The investigation of this paradigm on other IE tasks has not been studied yet. (2) We apply the common sentence embedding similarity for retrieving positive samples. We use self-consistency and confidence F1 score to obtain hard negative samples as in-context demonstrations. There might be other diverse strategies for measuring suitable positive samples and the quality of hard negative samples. (3) The performance of our work still lags behind previous fully-supervised methods. 




\bibliography{anthology,custom}

\clearpage
\appendix
\section{Implementation Experiment}
\label{sec:appendix_a}
\subsection{Dataset Statistics}
To ensure a comprehensive evaluation, we select a diverse set of datasets on the NER and RE tasks, 
including three NER benchmarks and four RE benchmarks. 
The specific statistics of these datasets, including the number of entity and relation types, as well as the distribution of instances across training, development, and test sets, are summarized in Table \ref{tab:datasets_anlasis}. 
This detailed breakdown provides insights into the dataset composition and serves as a reference for the robustness of the evaluation framework applied to our approach.
\begin{table}[ht]
\begin{adjustbox}{width=1\columnwidth,center}
\begin{tabular}{ccccccc}
\toprule
\multicolumn{2}{c}{Datasets}   & \tabincell{c}{Entity \\ Types}  & \tabincell{c}{Relation \\ Types} & Train & Dev  & Test \\ \hline
\multirow{3}{*}{NER} & CoNLL03 & 4            & /              & 14041 & 3250 & 3453 \\
                     & ACE04   & 7            & /              & 6202  & 745  & 812  \\
                     & ACE05-E   & 7            & /              & 7299  & 971  & 1060 \\ \hline
\multirow{4}{*}{RE}  & CoNLL04 & 4            & 5              & 922   & 231  & 288  \\
                     & ACE05   & 7            & 6              & 10051 & 2420 & 2050 \\
                     & NYT     & 3            & 24             & 56196 & 5000 & 5000 \\
                     & SciERC  & 6            & 7              & 1861  & 275  & 551 \\\bottomrule
\end{tabular}
\end{adjustbox}
\caption{Statistics of NER and RE Datasets.}
\label{tab:datasets_anlasis}
\vspace{-15pt}
\end{table}
\subsection{Implementation Experiment Details}
We run all experiments with the deep learning framework PyTorch NVIDIA Tesla A100 GPUs. 
The specific configurations and hyperparameters used in our method are meticulously chosen to optimize performance.  
These parameters include the maximum sequence length, the batch size, the number of beams for beam search, the top-p, and temperatures for controlling the randomness of the output.
The parameters are detailed in Table \ref{tab:parameters}.
\begin{table}[ht]
\begin{adjustbox}{width=0.8\columnwidth,center}
\begin{tabular}{lc}
\toprule
Parameters                        & Values   \\ \hline
Max Sequence Length               & 8192      \\
Batch Size                        & [1, 2]       \\
Num\_beams                        & 1         \\
Do\_sample                        & True      \\
Top\_p                            & 0.7       \\  
Temperature                       & [0.3,0.6,0.9]      \\ \bottomrule
\end{tabular}
\end{adjustbox}
\caption{The main parameters of our method \ourmethod{} based on CodeLlama.}
\label{tab:parameters}
\vspace{-15pt}
\end{table}
\section{Supplementary Analysis}
\label{sec:appendix_b}
\subsection{\ourmethod{} vs CodeIE based on CodeLlama}
Since ChatGPT \cite{openai-2022-chatgpt} is a robust model, our method, which is based on the open-source model CodeLlama, may be at a disadvantage. To better illustrate the effectiveness of our method \ourmethod{}, we reproduce CodeIE and replace its backbone with CodeLlama, and the relevant parameter settings are the same as ours. The results are shown in Table \ref{tab:codeie_rep_codellama_ner} and \ref{tab:codeie_rep_codellama_re}.
From the tables, under the same model and parameters, we can see that our method is superior to CodeIE. 
It shows that our method with wrong/negative samples can provide more effective information to large models to improve the performance of IE.
\begin{table}[ht]
\centering

    \centering
    \begin{adjustbox}{width=1\linewidth,center}
    \begin{tabular}{ccccc}
    \toprule
                                                  &                            & \multicolumn{3}{c}{NER}                                                                    \\
    \multirow{-2}{*}{Model}                       & \multirow{-2}{*}{Backbone} & \multicolumn{1}{c}{CoNLL03}           & \multicolumn{1}{c}{ACE04}    & \multicolumn{1}{c}{ACE05-E}  \\ \hline
                                                  & Code-davinci-002           & 82.32                                 & 55.29                        & 54.82                        \\
                                                  & CodeLlama-7B               & 72.33                                 & 36.21                        & 35.18 \\
                                                  & CodeLlama-13B              & 79.30                                 & 38.82                        & 35.87 \\
    \multirow{-4}{*}{CodeIE \cite{CodeIE-2023}}   & CodeLlama-34B              & 82.53                                 & 46.38                        & 46.46                            \\ \hline
                                                  & CodeLlama-7B               & 83.98                                 & 47.88                        & 45.65                        \\
                                                  & CodeLlama-13B              & 85.94                                 & 49.69                        & 48.04                        \\
    \multirow{-3}{*}{\textbf{c-ICL (ours)}}       & CodeLlama-34B              & 87.36                                 & 54.47                        & 55.65   \\ \bottomrule                    
    \end{tabular}
    \end{adjustbox}
    \caption{The experiment performances of \ourmethod{} and CodeIE via CodeLlama on NER benchmarks. }
    \label{tab:codeie_rep_codellama_ner}
    \vspace{-15pt}
\end{table}
\begin{table}[ht]
    \begin{adjustbox}{width=1\linewidth,center}
    \begin{tabular}{cccccc}
    \toprule
    \multirow{2}{*}{Model}                                     & \multirow{2}{*}{Backbone} & \multicolumn{4}{c}{RE}                                   \\
                                                               &                           & CoNLL04        & ACE05          & NYT            & SciERC         \\ \hline
    \multirow{4}{*}{CodeIE \cite{CodeIE-2023}}                 & Code-davinci-002          & 53.10          & 14.02          & 32.17          & 7.74           \\
                                                               & CodeLlama-7B              & 29.43          & 7.89           & 29.04          & 9.64           \\
                                                               & CodeLlama-13B             & 33.91          & 8.11           & 31.75          & 12.96        \\
                                                               & CodeLlama-34B             & 36.03          & 15.94          & 34.92               & 10.55          \\ \hline
    \multicolumn{1}{l}{\multirow{3}{*}{\textbf{c-ICL (ours)}}} & CodeLlama-7B              & 53.27          & 18.75          & 59.68          & 12.13 \\
    \multicolumn{1}{l}{}                                       & CodeLlama-13B             & 56.43          & 20.57          & 60.16          & 15.29 \\
    \multicolumn{1}{l}{}                                       & CodeLlama-34B             & 56.93          & 23.49          & 60.92          & 17.33 \\ \bottomrule        
    \end{tabular}
    \end{adjustbox}
    \caption{The experiment performances of \ourmethod{} and CodeIE via CodeLlama on RE benchmarks.}
    \label{tab:codeie_rep_codellama_re}
    \vspace{-15pt}
\end{table}
\subsection{Prediction Error}
In order to explore the impact of wrong/negative samples as in-context demonstrations on the types of errors that may occur in information extraction, we conducted analysis experiments on CoNLL03, ACE04, CoNLL04, and SciERC datasets via the backbone CodeLlama-7b. The results are shown in the Table \ref{tab:errors_ner} and \ref{tab:errors_re}. 
The number of entities involved in the evaluation of CoNLL03 and ACE04 datasets are 5648 and 3035 respectively.
The number of relations involved in the evaluation of CoNLL04 and SciERC datasets are 422 and 974 respectively.

\begin{table*}[ht]
\begin{adjustbox}{width=0.8\linewidth,center}
\begin{tabular}{lccccc}
\toprule
\multicolumn{1}{c}{\multirow{2}{*}{Model}} & \multirow{2}{*}{Backbone} & \multicolumn{2}{c}{Entity Type Error}  & \multicolumn{2}{c}{Entity Span Error} \\ \cline{3-6} 
\multicolumn{1}{c}{}                       &                           & CoNLL03                        & ACE04 & CoNLL03                       & ACE04 \\ \hline
CodeIE \cite{CodeIE-2023}                  & CodeLlama-7B              & 210                            &329     &2300                               &2303       \\ \hline
\textbf{c-ICL (ours)}                      & CodeLlama-7B              & 126                            & 434    &1842  &2549       \\ \hline
\end{tabular}
\end{adjustbox}
\caption{Predicton errors on NER datasets. "Ent Type Error" means the predicted entity type of the entity is not in the predefined type set. "Ent Span Error" means the predicted entity span of the entity is not in the test text.}
\label{tab:errors_ner}
\end{table*}

\begin{table*}[ht]
\begin{adjustbox}{width=1\linewidth,center}
\begin{tabular}{lccccccc}
\toprule
\multicolumn{1}{c}{\multirow{2}{*}{Model}} & \multirow{2}{*}{Backbone} & \multicolumn{2}{c}{Entity Type Error} & \multicolumn{2}{c}{Entity Span Error} & \multicolumn{2}{c}{Relation Type Error} \\ \cline{3-8} 
\multicolumn{1}{c}{}                       &                           & CoNLL04            & SciERC           & CoNLL04            & SciERC           & CoNLL04             & SciERC            \\ \hline
CodeIE \cite{CodeIE-2023}                  & CodeLlama-7B              & 13                 & 93               & 492                & 1481             & 16                  & 38                \\
\textbf{c-ICL   (ours)}                    & CodeLlama-7B              & 7                  & 88               & 273                & 1233             & 7                   & 17   \\ \bottomrule                 
\end{tabular}
\end{adjustbox}
\caption{Predicton errors on RE datasets. "Ent Type Error" means the predicted entity type of the entity is not in the predefined type set. "Ent Span Error" means the predicted span of the entity is not in the test text. "Relation Type Error" means the predicted label is not in the predefined relation type set.}
\label{tab:errors_re}
\vspace{-15pt}
\end{table*}

\section{Supplementary Case Study}
\label{sec:appendix_c}
\begin{figure*}[ht]
\begin{center}
    \subfigure[Example 2]{
        \includegraphics[width=1\columnwidth]{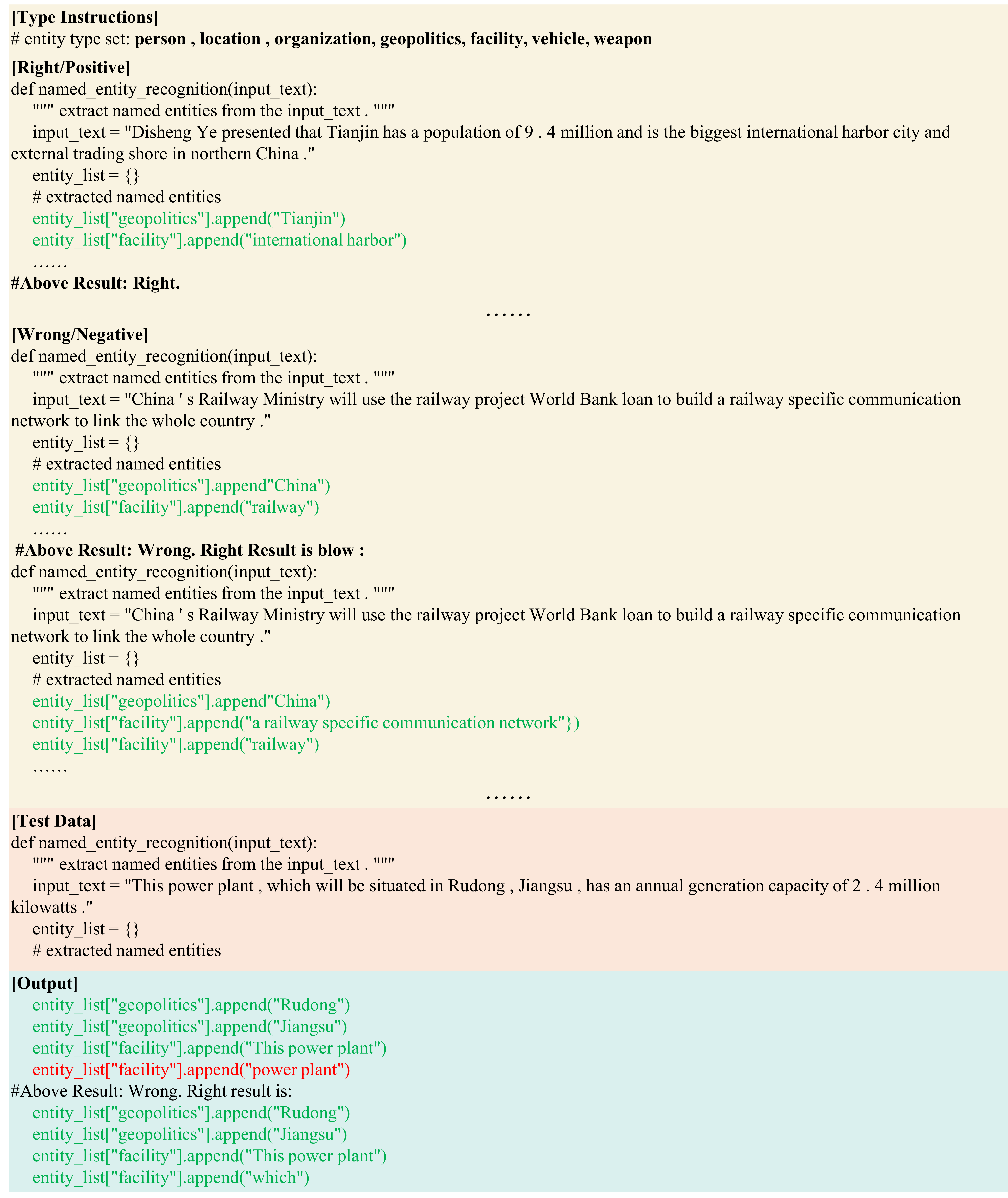}
        \label{Fig.case_study_2}    
    }
    \subfigure[Example 3]{
        \includegraphics[width=1\columnwidth]{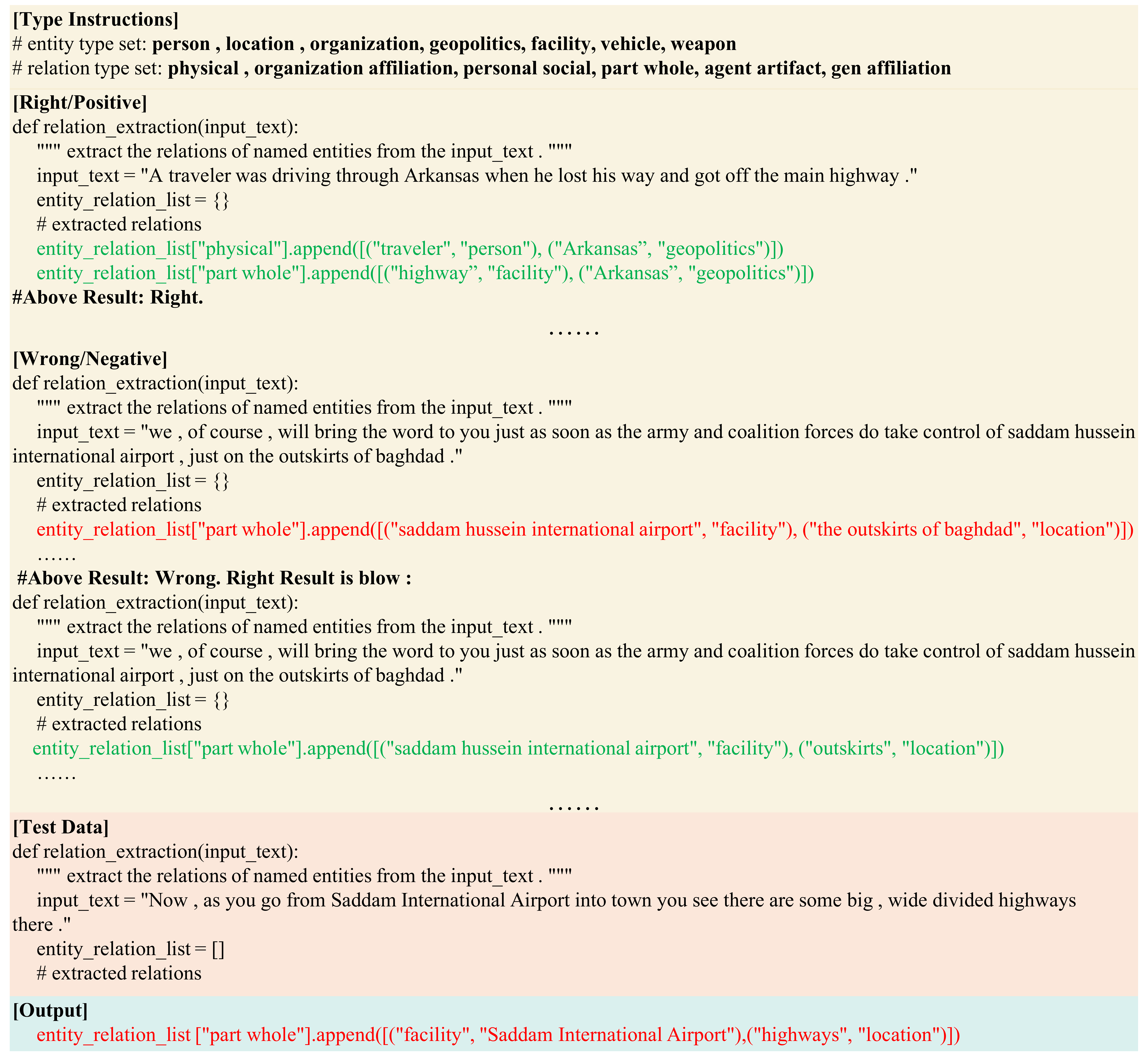} 
        \label{Fig.case_study_3}   
    }
    \caption{Supplementary case study of contrastive in-context learning. The red text means the incorrect labels and the green text means the right labels. Figure \ref{Fig.case_study_2} means the results of NER. Figure \ref{Fig.case_study_3} means the results of RE. }
    \label{Fig.other_case_study}
    \vspace{-15pt}
\end{center}
\end{figure*}

In this section, we present other examples of NER and RE test datasets in our experiments, 
as shown in Figure \ref{Fig.other_case_study}
In example 2, common issues that arise include the omission of entities or relations. 
By providing hints through negative samples, it is ensured that problems occurring in the generation process of test samples can be revised and corrected. 
In Example 3, even with the provision of semantically similar positive and negative samples as ICL demonstrations, the correct output results are still not guaranteed. 
In this case, it can be observed that the semantics of the test data itself are challenging.
``there'' means a location and ``highway'' is a facility in the location ``there''. 
The large model does not yet have handled perfectly to recognize such issues, even with the guidance provided by rich ICL demonstrations.
For this situation, the need for further advancements in training and fine-tuning techniques to improve LLM's interpretive capabilities and proficiency in handling complex inferences.
\end{document}